\documentclass{article} % For LaTeX2e
\usepackage{main,times}

\usepackage{amsmath,amsfonts,bm}

\def\eqref#1{equation~\ref{#1}}
\def\1{\bm{1}}

\DeclareMathAlphabet{\mathsfit}{\encodingdefault}{\sfdefault}{m}{sl}
\SetMathAlphabet{\mathsfit}{bold}{\encodingdefault}{\sfdefault}{bx}{n}

\newtheorem{definition}{Definition}

\usepackage{hyperref}
\usepackage{url}
\usepackage[capitalize,noabbrev]{cleveref}

\usepackage{tcolorbox}
\usepackage{booktabs}
\usepackage{multirow}
\usepackage{subcaption}
\usepackage{float}

\title{SCOPED:\\Score–Curvature Out-of-distribution\\ Proximity Evaluator for Diffusion}

\author{Brett Barkley \\
Department of Electrical and Computer Engineering\\
The University of Texas at Austin\\
Austin, TX, USA \\
\texttt{bbarkley@utexas.edu} \\
\And
Preston Culbertson \\
Department of Computer Science\\
Cornell University\\
Ithaca, NY, USA \\
\texttt{pculbertson@cornell.edu} \\
\And
David Fridovich-Keil \\
Department of Aerospace Engineering and Engineering Mechanics\\
The University of Texas at Austin\\
Austin, TX, USA \\
\texttt{dfk@utexas.edu} \\
}

\iclrfinalcopy % Uncomment for camera-ready version, but NOT for submission.
\begin{document}

\maketitle

\begin{abstract}
Out-of-distribution (OOD) detection is essential for reliable deployment of machine learning systems in vision, robotics,  reinforcement learning, and beyond. We introduce Score–Curvature Out-of-distribution Proximity Evaluator for Diffusion (SCOPED), a fast and general-purpose OOD detection method for diffusion models that reduces the number of forward passes on the trained model by an order of magnitude compared to prior methods, outperforming most diffusion-based baselines and closely approaching the accuracy of the strongest ones. SCOPED is computed from a single diffusion model trained once on a diverse dataset, and combines the Jacobian trace and squared norm of the model’s score function into a single test statistic. Rather than thresholding on a fixed value, we estimate the in-distribution density of SCOPED scores using kernel density estimation, enabling a flexible, unsupervised test that, in the simplest case, only requires a single forward pass and one Jacobian–vector product (JVP), made efficient by Hutchinson’s trace estimator. On four vision benchmarks, SCOPED achieves competitive or state-of-the-art precision-recall scores despite its low computational cost. The same method generalizes to robotic control tasks with shared state and action spaces, identifying distribution shifts across reward functions and training regimes. These results position SCOPED as a practical foundation for fast and reliable OOD detection in real-world domains, including perceptual artifacts in vision, outlier detection in autoregressive models, exploration in reinforcement learning, and dataset curation for unsupervised training.
\end{abstract}

\section{Introduction}
Out-of-distribution (OOD) detection is essential for the reliable deployment of machine learning systems in domains such as vision, robotics, and reinforcement learning. Modern models are prone to high confidence on abnormal or irrelevant inputs, leading to safety and robustness risks in real-world use \citep{nguyen2015deep, nalisnick2018deep, nalisnick2019detecting}. To address this, a large body of work has explored OOD detection methods \citep{hendrycks2018deep, liu2020energy, graham2023denoising, heng2024out}. Unsupervised methods are especially attractive since they require only in-distribution data, making them broadly applicable regardless of the eventual downstream task.

Generative modeling has emerged as a natural tool for OOD detection, but existing methods have key limitations. Likelihood-based approaches suffer from well-documented pathologies, such as assigning higher likelihood to OOD datasets than to the training set \citep{nalisnick2018deep, choi2018waic}. Reconstruction-based methods, including autoencoders and diffusion reconstructions, depend on carefully tuned information bottlenecks and are brittle in practice \citep{An2015VariationalAB, pinaya2021unsupervised}. More recent diffusion-based approaches such as DiffPath \citep{heng2024out} leverage trajectory geometry for OOD detection, but require evaluating the model along entire denoising paths which, like earlier diffusion-based OOD methods that require many model evaluations, is computationally expensive and poses a serious challenge for real-time and compute-limited applications. 

To address these shortcomings we propose the Score–Curvature Out-of-distribution Proximity Evaluator for Diffusion (SCOPED), a general-purpose alternative that requires far fewer model evaluations. Our method exploits fundamental intuition from information geometry \citep{cover2006elements}: near the typical set of a distribution (i.e., for ``in-distribution'' samples), the local curvature of the log-probability density is related to the norm of the score function. Since diffusion models learn the score directly \citep{ho2020denoising, song2021scorebased}, this curvature information can be obtained efficiently by a single Jacobian–vector product (JVP) through the trained model, provided the diffusion step is chosen appropriately. SCOPED constructs a simple statistic that summarizes the balance between curvature and score function norm, which serves as a reliable signal for whether a given query point is in- or out-of-distribution. Computing this statistic requires an order of magnitude fewer evaluations than the 10–1000 calls typical of prior diffusion-based OOD detection methods \citep{mahmood2020multiscale, heng2024out}. In addition, we introduce a simple offline procedure for selecting the diffusion step at which SCOPED is applied. Guided by the monotone decay of the signal-to-noise ratio under the forward diffusion process, we compute the SCOPED test statistic at a step early- and mid-way through the denoising schedule (chosen using only in-distribution data), ensuring robustness without any tuning on OOD datasets.

We first validate SCOPED in reinforcement learning, a setting where the notion of in-distribution vs. out-of-distribution data is less clear-cut than in vision. Environments can share identical dynamics yet differ in rewards (e.g., \texttt{reacher-easy} vs.\ \texttt{reacher-hard}, \texttt{humanoid-stand} vs.\ \texttt{humanoid-walk}), and independently trained policies can induce distinct state–action distributions even within the same task. On the DeepMind Control Suite (DMC) \citep{tassa2018deepmind} and D4RL Gym benchmarks \citep{fu2020d4rl}, SCOPED successfully separates such distribution shifts in proprioceptive RL domains. Detailed comparisons, including the effects of dataset diversity and replay mixtures, are provided in Section~\ref{sec:rl_results}.

We then turn to vision, where standardized baselines enable rigorous comparison. Across four benchmarks (CIFAR-10, SVHN, CelebA, CIFAR-100), SCOPED achieves competitive or state-of-the-art Area Under the Receiver Operating Characteristic curve (AUROC) scores while requiring an order of magnitude fewer diffusion model evaluations than most prior methods. Importantly, we use the same base diffusion model as the most comparable prior work \citep{heng2024out}, making clear that our efficiency gains are algorithmic rather than a consequence of dataset or model differences.

\begin{tcolorbox}[leftrule=1.5mm,top=1mm,bottom=0mm] 
\textbf{Key Contributions:}
\begin{enumerate}
    \item We introduce SCOPED, an OOD detection method that leverages information-theoretic connections between score norm and curvature to define a single test statistic. 
    SCOPED requires an order of magnitude fewer evaluations than competing diffusion-based methods, and --- unlike path-based approaches --- its probes are independent and can be fully parallelized across samples and timesteps. 
    We also propose a simple offline noise-level selection strategy using only in-distribution data.
    \item We provide a comprehensive evaluation of SCOPED across both proprioceptive reinforcement learning settings (DMC, D4RL) and vision benchmarks (CIFAR-10, SVHN, CelebA, CIFAR-100), showing that dataset choice fundamentally shapes OOD separability and that SCOPED achieves competitive or state-of-the-art AUROC on vision tasks.
\end{enumerate}
\end{tcolorbox}

\section{Background}
Our method applies to both discrete-time Denoising Diffusion Probabilistic Models (DDPMs) 
\citep{ho2020denoising} and continuous-time Elucidated Diffusion Models (EDMs) 
\citep{karras2022elucidating}, which are two widely used and mathematically connected 
formulations of score-based diffusion. 

\subsection{Score-Based Diffusion Models}
Diffusion models \citep{sohl2015deep, song2021scorebased} train a neural network to approximate 
the score function $s_\theta(x_t,t) \approx \nabla_x \log p_t(x_t)$ of progressively noised data $x_t$. 
At step $t$, Gaussian noise with variance $\sigma_t^2$ is added according to a schedule that grows with $t$, defining a distribution over noise scales from which training samples are drawn so the network learns to denoise across the full noise range. Given a noised input, the network predicts either the score or an equivalent representation (e.g. denoised data or noise), from which the score can be recovered; see \cref{app:diffusion_background} for details. Sampling then integrates the corresponding reverse dynamics to map Gaussian noise back into data. Early noise levels preserve fine detail, while later ones retain only coarse structure.

DDPM and EDM are discrete and continuous instances of this shared framework: both ultimately train a 
network to approximate $\nabla_x \log p_t(x_t)$. 
This common structure underlies our OOD detection method.
For completeness, we provide a more detailed derivation of DDPM, denoising score matching, and EDM in \cref{app:diffusion_background}.

\subsection{Typicality as a Measure of OOD}
\label{sec:ts}
A core idea in information theory is that \emph{typical} outcomes of a random process are not necessarily the most probable. Typical sets are classically defined in terms of sequences:
a sequence of i.i.d. draws $\{x_1, x_2, \dots, x_n\} \sim p$ is $\delta$-typical if 
$$H(p) - \delta \le -\frac{1}{n}\log p(x_1, x_2, \dots, x_n) \le H(p) + \delta\,,$$
where $H(p) = -\mathbb{E}_p[\log p(x)]$ is the (differential) entropy of density $p$. The asymptotic equipartition property (AEP) ensures that, in the limit $n \to \infty$, any sequence $\{x_i\}_{i=1}^n$ will be $0$-typical with probability 1 \citep{cover2006elements}. 

Although the typical set is defined in terms of i.i.d. sequences, it is also natural to discuss typicality for length $n=1$ sequences of high-dimensional distributions.
Consider the example of an isotropic Gaussian distribution in high dimension $d$, where the density is maximized at the mean, yet almost all mass lies in a thin shell with radius $\sqrt{d}$. This counterintuitive phenomenon, sometimes referred to as the Gaussian annulus theorem \citep{vershynin2018high, nalisnick2019detecting}, tells us that with high probability any random draw will lie on the shell, rather than near the mode at the origin. This shell is precisely the set of points at which $-\log p(x) = H(p)$. 

This perspective has two consequences for OOD detection. First, it explains why likelihood-based methods can fail: points with high density may be atypical if they fall outside the shell where mass concentrates. Second, it suggests that in-distribution data should lie in the typical set. Formally, then, we define OOD points as follows.

\begin{definition}[$\delta$-OOD points]
    \label{def:ood}
    We call a sequence of $n$ points, $\{x_1, x_2, \dots, x_n\} \overset{\text{i.i.d.}}{\sim} p$, $\delta$-OOD for distribution $p$ if they are not $\delta$-typical.
\end{definition}

In practice, we seek to identify length $n=1$ $\delta$-OOD sequences, with $\delta$ specified indirectly by comparing the test statistic to an adaptive threshold calibrated on in-distribution data.

\subsection{The Score-Curvature Ratio}

Let $p(x)$ be a nondegenerate density and define the score $s(x) := -\nabla_x \log p(x)$. Under standard regularity conditions, classical results provide two equivalent expressions for the Fisher information:
\[
\mathbb{E}_p\!\left[\nabla_{x}^2 \log p(x)\right] \;=\; -\,\mathbb{E}_p\!\left[s(x)\,s(x)^\top\right],
\qquad
\text{hence}\qquad
\mathbb{E}_p\!\left[\operatorname{Tr}\big(\nabla_x s(x)\big)\right] \;=\; \mathbb{E}_p\!\left[\|s(x)\|^2\right],
\]
cf. \citep{Gallager_2013}. 
The local curvature
\[
\kappa(x) \;:=\; \operatorname{Tr}\big(\nabla_x s(x)\big) \;=\; -\,\operatorname{Tr}\big(\nabla_x^2 \log p(x)\big)
\]
is nonnegative for log-concave models and positive for Gaussians. 
For in-distribution samples drawn from $p$, therefore, the expected score magnitude (measured by $p$) matches expected curvature. 

Nevertheless, one can still define the statistic
\[
T(x) \;=\; \frac{\|s(x)\|^2}{\kappa(x)}
\] 
for any point $x$ at which $\kappa(x) \ne 0$, with the intuition that high-dimensional 
distributions will assign virtually all probability mass to typical samples, with 
$-\log p(x) \approx H(p)$ and at which we can expect $T(x) \approx 1$. For numerical stability, when computing $T(x)$
we add a small $\varepsilon$ to the denominator to avoid division by zero, though in practice we observe
that $\kappa(x)$ always takes on large values.

As a concrete example, consider the $d$-dimensional isotropic Gaussian setting: $x \sim \mathcal N(0,\sigma^2 I_d)$. In this case the score is $s(x)=x/\sigma^2$ and the curvature is $\kappa(x)=d/\sigma^2$, so
\[
T(x) = \frac{\|s(x)\|^2}{\kappa(x)} = \frac{\|x\|^2}{d\,\sigma^2}.
\]
Because $\|x\|^2$ concentrates sharply around $d\sigma^2$ in high dimensions, $T(x)$ remains close to one on the typical set, while samples far from this shell yield values that deviate substantially. This intuition forms the core of our approach: \emph{we hypothesize that the statistic $T(x)$ is highly predictive of a point's membership in the $\delta$-typical set, and propose to use it as a metric for OOD detection.}

Importantly, diffusion models provide direct access to both the score $s(x)$ and the curvature $\kappa(x)$ (via the Jacobian trace, which can be estimated efficiently). This makes $T(x)$ computable from a single trained diffusion model at a chosen noise level, thereby linking the information-theoretic notion of the typical set to a practical, model-based criterion for out-of-distribution detection. 

A final subtlety is that the typicality ratio admits a global sign ambiguity: the squared norm in the numerator discards orientation, while the Jacobian trace in the denominator retains sign information. We correct this with a global sign factor, as detailed in \cref{app:sign}, and confirm in \cref{sec:signambig} that this correction is necessary for stable OOD performance in vision-based tasks.

\section{From the Score-Curvature Ratio to a Practical OOD Metric: SCOPED}  

\subsection{Achieving Computational Efficiency}
A key advantage of SCOPED over path-based diffusion approaches such as DiffPath \citep{heng2024out} is its independence from serial trajectory integration. DiffPath requires propagating samples through the probability flow ordinary differential equation (ODE) with Denoising Diffusion Implicit Models (DDIM) \citep{song2021scorebased}, where each step depends on the previous one. This makes the method inherently sequential: the cost scales linearly with the number of integration steps, and timesteps cannot be parallelized. As a result, the number of function evaluations (NFEs) grows directly with the trajectory length. 

In contrast, SCOPED probes the diffusion model at arbitrary noise levels without reconstructing the full denoising path. Each statistic is computed from a forward evaluation and a single JVP at a chosen timestep, independent of all other timesteps. This independence means that (i) NFEs are reduced by roughly an order of magnitude compared to path-based methods, since only a handful of strategically chosen timesteps are required, and (ii) those evaluations are not serially dependent and can therefore be fully parallelized across both samples and timesteps on modern accelerators. In practice, this combination of reduced counts and effective parallelism yields efficiency gains that exceed what raw NFE tallies alone would suggest.

A second source of efficiency comes from curvature estimation. Naively, evaluating the Jacobian trace $\operatorname{Tr}(\nabla_x s(x))$ is prohibitively expensive in high dimensions. We instead employ Hutchinson’s stochastic trace estimator~\citep{hutchinson1989stochastic}, which uses randomized projections to form an unbiased estimate with cost linear in the dimension. Each probe requires only a single JVP with the score network, avoiding explicit Jacobian matrix formation. In practice, a single probe is often sufficient, and averaging a few probes can further reduce variance~\citep{pearlmutter1994fast,martens2010deep}. This makes SCOPED at a given noise level comparable in cost to one additional forward pass of the score network, since Hutchinson’s estimator avoids building the full Jacobian.

Together, these properties make SCOPED practical at scale: an order of magnitude fewer NFEs than path-based diffusion methods, parallelism across timesteps and samples, and curvature estimation that avoids the quadratic cost of explicit Jacobian formation, requiring only a few JVPs.

\subsection{Motivation and OOD Detection with Kernel Density Estimation}
Although $T(x)$ measures typicality and is feasible to compute with Hutchinson's estimator, its raw values are not universally calibrated across datasets, diffusion noise levels, or models. On its own, $T(x)$ is therefore not a reliable discriminator of in-distribution (ID) versus out-of-distribution (OOD) data. We illustrate this with a simple RL example, then motivate our approach for calibration.

\begin{figure}[t]
\centering
\begin{subfigure}{0.35\textwidth}
    \includegraphics[width=\linewidth]{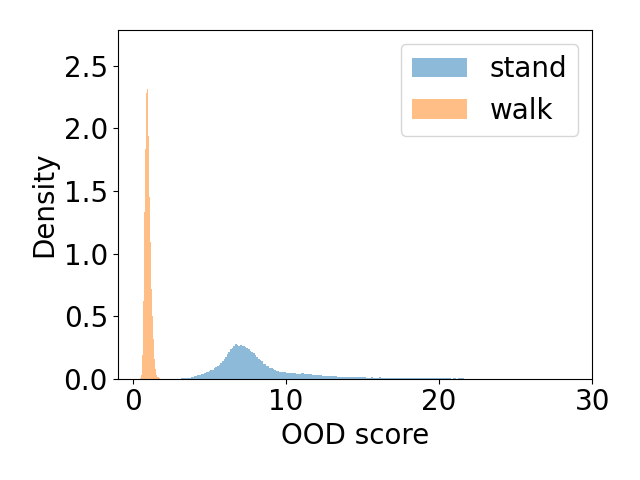}
    \caption{Trained on \texttt{humanoid-walk}}
\end{subfigure}
\begin{subfigure}{0.35\textwidth}
    \includegraphics[width=\linewidth]{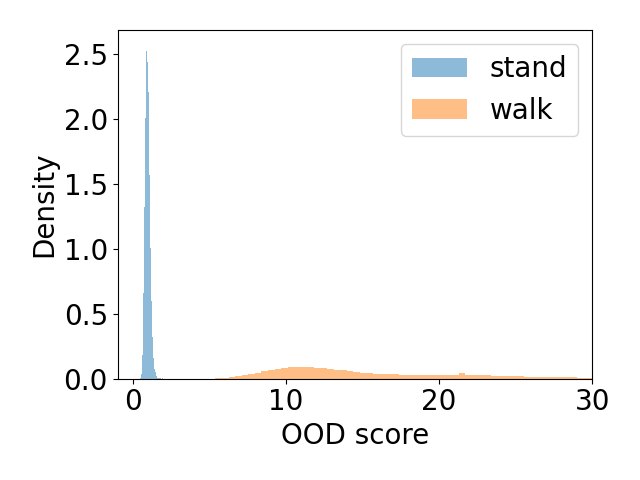}
    \caption{Trained on \texttt{humanoid-stand}}
\end{subfigure}
\caption{Distribution of SCOPED OOD scores $T(x)$ for \texttt{humanoid-stand} and \texttt{humanoid-walk} transitions. 
In-distribution evaluations concentrate near $1$ with tight quantiles, while cross-task evaluations shift upward and become more dispersed, indicating separability.}
\label{fig:humanoid_tx}
\end{figure}

\paragraph{Motivating example in reinforcement learning.}
We train high-proficiency Soft Actor-Critic (SAC) policies \citep{haarnoja_soft_2019} on \texttt{humanoid-stand} and \texttt{humanoid-walk}, which are high-dimensional, proprioceptive, continuous control robotics tasks from the DeepMind Control Suite \citep{tassa_dm_control_2020} with identical dynamics but non-overlapping rewards. Each trained policy is rolled out deterministically to create a transition replay buffer of observation-action-next observation-reward tuples %$(o,a,o',r)$
per task. We fit an Elucidated Diffusion Model (EDM) denoiser \citep{lu2023synthetic} to each dataset and evaluate the typicality ratio $T(x)$ on each transition in both ID and cross-task buffers. As shown in \cref{fig:humanoid_tx}, ID evaluations per transition concentrate near $1$ with tight quantiles, while cross-task evaluations shift upward and become much more dispersed. This pair of datasets is trivially separable via the $T(x)$ statistic, which serves as a sanity check that $T(x)$ can capture distributional differences; however, we will see below that care must be taken to choose an effective threshold when data is not trivially separable.

\paragraph{Calibration for a practical detector.}
Despite the separation in the above example, absolute $T(x)$ values are not always comparable across datasets and models, so a fixed threshold is unreliable. During training we have access to ID samples. We therefore compute $T(x)$ on ID data, and fit a density to these values using kernel density estimation (KDE), which estimates a smooth probability density by placing Gaussian kernels around each observed value. The result is a continuous estimate of the ID test statistic distribution without assuming a specific parametric form. This converts typicality into a calibrated anomaly score, given by $-\log h(T(x))$, where $h$ is the KDE fit on in-distribution values of $T(x)$, which we use for OOD detection with empirical results in \cref{sec:results}.

It is important to note, however, that the effectiveness of this calibration depends on the domain and the dataset used to train the diffusion model, which need not coincide with the ID data. In this RL example, we found that separability between \texttt{humanoid-stand} and \texttt{humanoid-walk} was robust to the choice of training dataset even without introducing the KDE concept. In contrast, as we show in \cref{sec:results} and mirroring prior work \citep{heng2024out}, training dataset choice plays a critical role in vision-based settings and even other proprioceptive datasets, where separability can be strongly affected by the diversity and coverage of the training distribution.

\subsection{Offline Selection of Noise Levels for SCOPED}
\label{sec:noise_selection}
Choosing the noise level is crucial for balancing retained signal with injected noise.  

For our proprioceptive D4RL and DMC tasks, we evaluate at a single noise level given by the mode of the log-normal prior over noise scales, $\sigma_{\text{mode}} = \exp(\mu - \sigma_{\log}^2)$, 
where $\mu$ and $\sigma_{\log}^2$ are the mean and variance of $\log \sigma$. This corresponds to the most probable noise scale under the prior, which lies toward the early-middle part of the schedule. It suppresses fine state-action detail while preserving coarse structure, providing a principled single-point evaluation without requiring a sweep.

In contrast, for our vision experiments, we estimate the signal-to-noise ratio (SNR) of the distribution $p_t(x_t)$ offline using in-distribution data (cf. \cref{app:snr_cifar10}), at different points along the forward diffusion process $t$. SNR decays monotonically: early timesteps preserve fine detail, while later ones approach noise dominance. To ensure robustness across this spectrum, we evaluate SCOPED at two points: an early step ($t=1$) that maximizes fine information, and a mid-level step ($t=300$) where about 95\% of the signal remains, the SNR curve enters a roughly linear decline, and coarse structure is retained while some fine detail is suppressed.

For each test input we compute anomaly scores (negative log-likelihoods under the KDE fit) at both timesteps and take their maximum as the final anomaly score. This is fully unsupervised, requires only ID data, and adds negligible cost since the KDE fit is precomputed offline and only the typicality ratio is evaluated online. Specifically, we fit a KDE to score–curvature statistics of ID data at the chosen timesteps and compute the negative log-likelihood of the typicality score for a test input under this density model. For evaluation we report AUROC, which is threshold-free, whereas for deployment a cutoff can be set from ID data.

We define two reference variants. SCOPED (Single) fixes the noise level at $t=300$, chosen once from in-distribution data using the same procedure as above. SCOPED (Oracle) assumes knowledge of the best-performing timestep for each ID/OOD dataset pair, providing an informative but unattainable upper bound.

We emphasize that, unlike prior diffusion-based methods such as DiffPath \citep{heng2024out}, which sweep hyperparameters on OOD benchmarks, SCOPED selects noise levels using only in-distribution data. Our ablations show robustness, not tuning.

\section{Related Work}
Out-of-distribution (OOD) detection has been studied through confidence scores \citep{hendrycks2018deep, liang2017enhancing}, generative likelihoods \citep{nalisnick2019detecting, choi2018waic, ren2019likelihood}, and reconstruction methods \citep{An2015VariationalAB, pinaya2021unsupervised}. While effective in narrow cases, these approaches often fail under dataset shifts, such as high likelihoods on OOD data or sensitivity of autoencoder reconstructions to bottleneck design.

Diffusion models have recently been applied to OOD detection through reconstruction \citep{graham2023denoising, liu2023unsupervised} and through score-based or trajectory-based statistics \citep{mahmood2020multiscale, heng2024out}. However, such approaches often rely on multiple evaluations across timesteps or full denoising paths, and remain limited by the broader challenge that training OOD detectors only on ID data is misaligned with the OOD objective \citep{li2025out}. SCOPED addresses these issues by deriving a single statistic from the score and its curvature at a chosen noise level, which reduces computation and grounds detection in generative model geometry. It can also incorporate a foundation dataset alongside ID data to mitigate the ID-only misalignment. To our knowledge, SCOPED is also among the first approaches to validate diffusion-based OOD detection on RL benchmarks, including D4RL and the DeepMind Control Suite \citep{fu2020d4rl, tassa2018deepmind}.

\section{Results}
\label{sec:results}

As in prior work, experiments here compare OOD detection across dataset pairs.

For RL, we generate transition tuples (observation, action, next observation, reward) from DeepMind Control Suite (DMC) tasks such as \texttt{humanoid-stand} and \texttt{humanoid-walk} \citep{tassa2018deepmind}, as well as OpenAI Gym tasks \citep{brockman2016openai} from the offline D4RL benchmark \citep{fu2020d4rl}. These cover a range of continuous control problems and allow us to examine shifts across reward functions and independently trained policies. Since OOD baselines are overwhelmingly vision-focused, we use these RL tasks primarily as sanity checks and case studies, not competitive benchmarks.

For vision-based tasks, we evaluate OOD detection on CIFAR-10 (C10), SVHN, CelebA, and CIFAR-100 (C100), which cover both $32\times 32$ and $64\times 64$ resolutions. We use Denoising Diffusion Probabilistic Models (DDPMs) \citep{ho2020denoising}, trained unconditionally on CelebA to match the strongest comparable baselines \citep{heng2024out}. Following \citet{heng2024out}, we adopt a consistent resizing procedure to avoid artifacts that simplify detection due to data fidelity discrepancies: when datasets have different original resolutions, higher-resolution images are first downsampled to match the lower-resolution set, and both are then upsampled to the model’s input resolution. This ensures consistent blur across datasets and prevents artificially inflated OOD detection performance. 

We benchmark performance in vision tasks against strong generative baselines, including DiffPath \citep{heng2024out}, IGEBM \citep{du2019implicit}, VAEBM \citep{xiao2020vaebm}, and Improved CD \citep{du2021improved}. We also include likelihood-driven tests, namely Input Complexity \citep{serra2019input}, Density of States \citep{morningstar2021density}, WAIC \citep{choi2018waic}, Typicality \citep{nalisnick2019detecting}, and the Likelihood Ratio \citep{ren2019likelihood} applied to Glow \citep{kingma2018glow}. Finally, we compare to diffusion-based baselines: vanilla NLL, Input Complexity computed from diffusion model likelihoods, and implementations of DDPM-OOD \citep{graham2023denoising}, LMD \citep{liu2023unsupervised}, and MSMA \citep{mahmood2020multiscale}. We report AUROC and computational cost measured as the number of function evaluations, defined as forward passes. For comparability, our method also performs one Jacobian–vector product per evaluation, and we note this cost in the totals.

\subsection{Reinforcement Learning Case Studies}
\label{sec:rl_results}
We first consider reinforcement learning, a domain where the definition of in-distribution is inherently ambiguous. Environments can share identical dynamics while differing in rewards, and even within a single task, different policies can induce markedly different state–action distributions due to the highly stochastic nature of RL training. To capture these axes of divergence, we evaluate SCOPED across three categories: reward changes (e.g., \texttt{reacher-easy} vs. \texttt{reacher-hard}, \texttt{humanoid-stand} vs. \texttt{humanoid-walk}), transitions induced by different behavior policies as collected in different replay buffer realizations (e.g., \texttt{ant-random} vs. \texttt{ant-expert}), and policy realizations that differ only by random seed (e.g., \texttt{reacher-hard} seed A vs. seed B).  

\paragraph{DMC results.}
On DeepMind Control Suite tasks, SCOPED achieves perfect separation across all tested shifts (\cref{app:offlinedmc}, Table~\ref{tab:dmc}). These results validate that the method captures distributional divergence not only in cases where the state–action–reward distributions are clearly non-overlapping, such as \texttt{humanoid-stand} versus \texttt{humanoid-walk} (cf. \cref{fig:humanoid_tx}), but also in settings where overlap is expected, such as \texttt{finger-turn} or \texttt{reacher} tasks. In these latter cases, the same dynamics and similar reward structures generate transitions that occupy overlapping regions of the state-action space, yet SCOPED still produces strong separation and perfect out-of-distribution detection accuracy. This outcome would not have been obvious a priori and suggests that the statistic is sensitive to more subtle distributional cues than reward specification alone. While these experiments may appear straightforward a posteriori, they serve as valuable sanity checks: they confirm that SCOPED behaves in line with intuition when distributions are clearly distinct, while also demonstrating sensitivity in cases where overlap exists and separability is less obvious. Together, they provide a foundation before turning to more challenging benchmarks. 

\begin{figure}[h]
    \centering
    \includegraphics[width=0.6\textwidth]{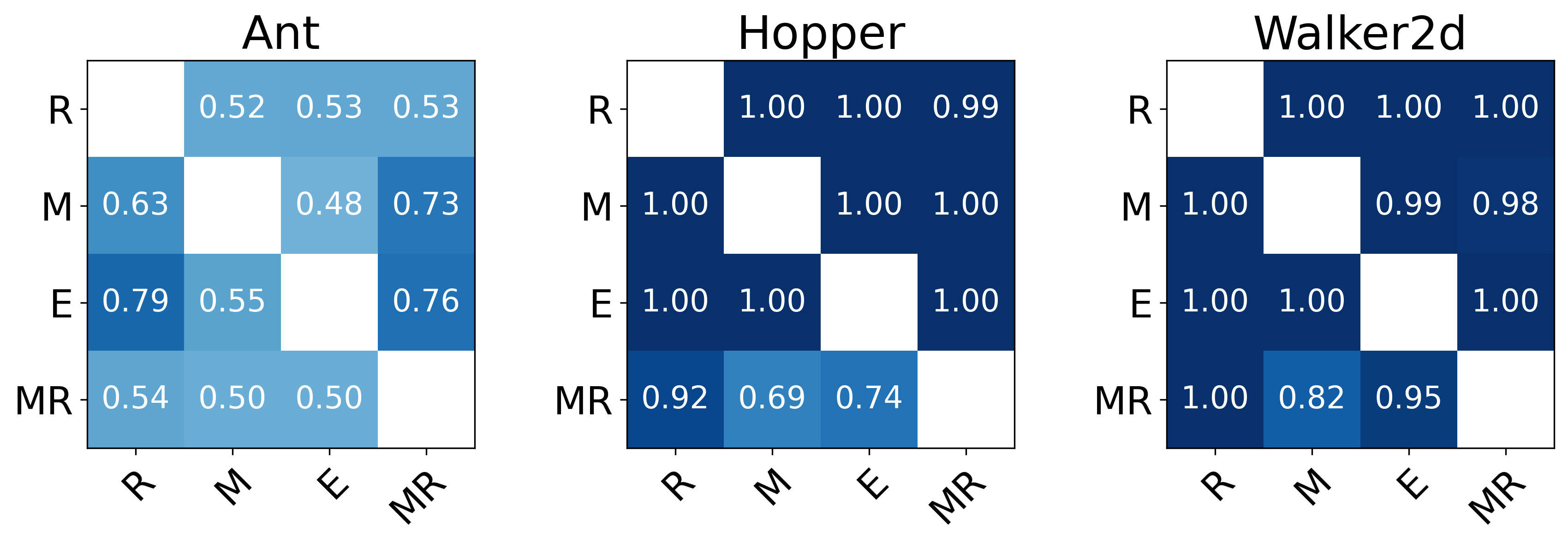}
    \caption{AUROC heatmaps for D4RL. Rows indicate the training buffer and columns the OOD buffer. D4RL data is abbreviated R (random), M (medium), E (expert), and MR (medium-replay).}
    \label{fig:d4rl_heatmaps}
\end{figure}

\paragraph{D4RL results.}
The D4RL benchmarks provide a more nuanced picture (cf. \cref{fig:d4rl_heatmaps}). Each dataset is collected from a different behavior policy \citep{fu2020d4rl}: \texttt{random} from an untrained agent, \texttt{expert} from a fully trained SAC policy, \texttt{medium} from a partially trained policy at about one-third expert performance, and \texttt{medium-replay} from the full replay buffer of the medium agent throughout training, which is more diverse but also noisier due to exploratory behavior.  

In \texttt{hopper-v2} and \texttt{walker2d-v2}, SCOPED achieves near-perfect AUROC across most pairs, while \texttt{ant-v2} is harder, especially for medium vs.\ expert or medium-replay. A key observation is that training on the most diverse dataset (\texttt{medium-replay}) does not produce the strongest detector: it consistently underperforms in Hopper and Walker, and only conditionally helps in Ant.  

Although counterintuitive, this result is consistent with vision-based OOD methods \citep{heng2024out} that rely on a single foundation model rather than ID training: performance depends heavily on the choice of training distribution, and more diversity of dataset coverage does not always help. In RL, termination conditions constrain random or partially trained agents to narrow regions of state space, while replay mixtures add noisy and inconsistent trajectories. This makes superficially diverse buffers more entangled and less separable. In contrast, proficient and diverse policies such as expert agents generate broad but coherent coverage, sharpening the boundary between in- and out-of-distribution as is consistently shown by our results. Dataset choice is therefore a central factor in RL OOD detection, and intuitions from vision that “more coverage helps” do not transfer directly.

\paragraph{Interpretation.}
Taken together, these case studies demonstrate that SCOPED naturally extends to proprioceptive RL data while revealing important trade-offs. The DMC results serve as clear sanity checks, while the D4RL results illustrate more subtle distinctions that, for example, configurations that maximize separability for one dataset pair may perform poorly for another. In particular, termination conditions and replay mixtures can make diverse buffers less separable, whereas expert datasets sharpen boundaries and yield stronger detectors. We emphasize that these studies are not intended as a competitive benchmark, but as diagnostic explorations of how SCOPED interacts with different notions of distribution shift in RL --- reward changes, policy changes, and distinct policy realizations. With this foundation, we next turn to vision benchmarks, where standardized datasets and strong baselines allow more rigorous quantitative comparison.

\subsection{Main Vision-Based Results}

Before turning to results, we first clarify the setup for vision-based OOD detection.

\paragraph{Foundation dataset.}
\label{para:foundation}
We follow \citet{heng2024out} and use an unconditional DDPM trained on CelebA (available \href{https://huggingface.co/ajrheng/diffpath/tree/main}{here}) as the backbone for our diffusion OOD detector. CelebA is a natural choice because it is large and diverse, and has been shown to yield strong performance in prior diffusion-based OOD work. Using the same model also ensures that our results are directly comparable to the strongest diffusion-based baselines, without confounding effects from architecture or training differences. In this setting CelebA serves as the \emph{foundation dataset} for model training, while ID/OOD evaluation is carried out across CIFAR-10, SVHN, CelebA, and CIFAR-100.

\paragraph{Results.}
Table~\ref{tab:vision_results} reports AUROC across CIFAR-10, SVHN, CelebA, and CIFAR-100, along with the number of function evaluations (NFE), conventionally measured as the number of forward passes through the network. To make accounting explicit, we denote a forward network evaluation as \texttt{1F} and a JVP as \texttt{1J}. Although the precise complexity of these operations depends on architecture, implementation, and batch size, prior work has noted that a JVP has equivalent cost to approximately one forward \citep{baydin2018automatic}, so we adopt the convention that \texttt{1J} $\approx$ \texttt{1F}. Each SCOPED evaluation therefore requires \texttt{1F}+\texttt{1J} (two effective forward passes), while the two-step variant uses \texttt{2F}+\texttt{2J}. Although this at minimum doubles the nominal cost relative to forward-only methods, the crucial difference is that SCOPED evaluations are independent and can be fully parallelized across samples and timesteps. By contrast, trajectory-based methods such as DiffPath require serial integration steps whose cost scales linearly with path length. In practice, this independence means SCOPED achieves substantially lower wall-clock time despite the need for a JVP.

With this setup, the results show that SCOPED achieves competitive or state-of-the-art detection accuracy while requiring orders of magnitude fewer effective function evaluations than prior methods. The default two-step variant (\texttt{2F}+\texttt{2J}) offers robustness across dataset pairs while remaining far cheaper than trajectory-based approaches. Ablations over different early–mid step pairs (\cref{sec:ablation_2step}) confirm that AUROC remains consistently high across choices, demonstrating that the two-step procedure provides robust performance across timesteps.

We also evaluate SCOPED (Single), which fixes the noise level at $t=300$ chosen once from in-distribution data. By omitting the earlier fine-detail step, this variant sacrifices some robustness, %since this mid-level choice omits fine-grained details, 
yet it demonstrates that strong performance is achievable with only \texttt{1F}+\texttt{1J}. Ablations varying the mid-step choice (\cref{sec:ablation_1step}) confirm that performance remains relatively insensitive to the exact choice of step.

Finally, SCOPED (Oracle) assumes knowledge of the best-performing timestep for each ID/OOD dataset pair, constructed by selecting the single step that yields the highest AUROC (highlighted in bold in \cref{tab:onestep}). While unavailable in practice, it provides an informative upper bound. Importantly, all noise levels for SCOPED and SCOPED (Single) are selected using only in-distribution data; unlike some prior diffusion-based OOD methods, our ablations do not tune against OOD benchmarks, preventing evaluation leakage.

\begin{table}[h!]
\begin{center}
\scriptsize
\caption{
AUROC scores for in- vs out-of-distribution tasks. Higher is better. 
\underline{Underline} second best, \textbf{bold} best (Oracle excluded).
Because SCOPED evaluations are independent across samples and timesteps, they parallelize fully, making wall-clock cost substantially lower than path-based methods despite the additional JVPs. 
We report computational cost as \#\texttt{F} + \#\texttt{J}, where \texttt{F} denotes a forward pass and \texttt{J} a Jacobian–vector product. 
Baseline results reproduced from \citet{heng2024out}.
}
\resizebox{\textwidth}{!}{
\begin{tabular}{lccccccccccc}
\toprule
\textbf{Method} & \multicolumn{3}{c}{\textbf{C10 vs}} & \multicolumn{3}{c}{\textbf{SVHN vs}} & \multicolumn{3}{c}{\textbf{CelebA vs}} & \textbf{Avg} & \textbf{\#\texttt{F} + \#\texttt{J}} \\
 & SVHN & CelebA & C100 & C10 & CelebA & C100 & C10 & SVHN & C100 & & \\
\cmidrule(lr){2-4} \cmidrule(lr){5-7} \cmidrule(lr){8-10}
IC & 0.950 & 0.863 & \underline{0.736} & - & - & - & - & - & - & - & - \\
IGEBM & 0.630 & 0.700 & 0.500 & - & - & - & - & - & - & - & - \\
VAEBM & 0.830 & 0.770 & 0.620 & - & - & - & - & - & - & - & - \\
Improved CD & 0.910 & -     & \textbf{0.830} & - & - & - & - & - & - & - & - \\
DoS & 0.955 & \underline{0.995} & 0.571 & 0.962 & \textbf{1.00} & 0.965 & 0.949 & \underline{0.997} & 0.956 & \textbf{0.928} & - \\
WAIC$^1$ & 0.143 & 0.928 & 0.532 & 0.802 & 0.991 & 0.831 & 0.507 & 0.139 & 0.535 & 0.601 & - \\
TT$^1$ & 0.870 & 0.848 & 0.548 & 0.970 & \textbf{1.00} & 0.965 & 0.634 & 0.982 & 0.671 & 0.832 & - \\
LR$^1$ & 0.064 & 0.914 & 0.520 & 0.819 & 0.912 & 0.779 & 0.323 & 0.028 & 0.357 & 0.524 & - \\
\midrule
\multicolumn{12}{c}{\textbf{Diffusion-based}} \\
\midrule
NLL & 0.091 & 0.574 & 0.521 & \textbf{0.990} & \underline{0.999} & \textbf{0.992} & 0.814 & 0.105 & 0.786 & 0.652 & 1000\texttt{F} + 0\texttt{J} \\
IC & 0.921 & 0.516 & 0.519 & 0.080 & 0.028 & 0.100 & 0.485 & 0.972 & 0.510 & 0.459 & 1000\texttt{F} + 0\texttt{J} \\
MSMA & \underline{0.957} & \textbf{1.00} & 0.615 & \underline{0.976} & 0.995 & \underline{0.980} & 0.910 & 0.996 & 0.927 & \textbf{0.928} & 10\texttt{F} + 0\texttt{J} \\
DDPM-OOD & 0.390 & 0.659 & 0.536 & 0.951 & 0.986 & 0.945 & 0.795 & 0.636 & 0.778 & 0.742 & 350\texttt{F} + 0\texttt{J} \\
LMD & \textbf{0.992} & 0.557 & 0.604 & 0.919 & 0.890 & 0.881 & \underline{0.989} & \textbf{1.00} & \underline{0.979} & 0.868 & $10^4$\texttt{F} + 0\texttt{J} \\
\midrule
\multicolumn{12}{c}{\textbf{Curvature and Diffusion-Based}} \\
\midrule
DiffPath & 0.910 & 0.897 & 0.590 & 0.939 & 0.979 & 0.953 & \textbf{0.998} & \textbf{1.00} & \textbf{0.998} & \underline{0.918} & 10\texttt{F} + 0\texttt{J} \\
SCOPED & 0.814 & 0.940 & 0.477 & 0.971 & 0.996 & 0.959 & 0.925 & 0.994 & 0.962 & 0.892 & \underline{2\texttt{F} + 2\texttt{J}} \\
SCOPED (Single) & 0.774 & 0.867 & 0.460 & \underline{0.976} & \textbf{1.00} & 0.966 & 0.938 & \textbf{1.00} & 0.971 & 0.884 & \textbf{1\texttt{F} + 1\texttt{J}} \\
\midrule
SCOPED (Oracle) & 0.964 & 0.885 & 0.768 & 0.993 & 1.00 & 0.975 & 0.938 & 1.00 & 0.971 & 0.944 & 1\texttt{F} + 1\texttt{J} \\
\bottomrule
\end{tabular}
}
\label{tab:vision_results}
\end{center}
\end{table}

\section{Conclusion}
In this work we introduced SCOPED, a diffusion-based OOD detection method that leverages the score–curvature ratio as a simple, theoretically motivated, and computationally efficient test statistic that is highly parallelizable. Unlike prior approaches that require long diffusion trajectories or many function evaluations, SCOPED achieves competitive or state-of-the-art performance on both vision and RL benchmarks while reducing model evaluations by an order of magnitude.

Our experiments demonstrate that SCOPED generalizes across domains, cleanly separating distribution shifts in RL and delivering strong accuracy on standardized vision benchmarks. These results show that efficiency and robustness need not be traded off, and that simple geometric properties of the score function provide powerful signals for detecting atypical inputs.

Looking ahead, the success of SCOPED suggests a broader design principle: OOD detection can be grounded in the geometric statistics of generative models rather than expensive likelihood or path-based estimators. Future work could combine SCOPED with more advanced noise-step selection methods, integrate it with exploration strategies in RL, or extend it to autoregressive models and multimodal domains. We hope that SCOPED provides both a practical tool for fast OOD detection and a conceptual bridge between information-theoretic typicality and generative model geometry.

\section{Acknowledgments}
This work was supported by the National Science Foundation under Grant No. 2409535.

\newpage

\bibliography{main}
\bibliographystyle{main}

\appendix

\section{Usage of Large Language Models (LLMs)}
LLMs were used for paper writing assistance and to aid in brainstorming, software development, and experiment design.

\section{Reproducibility Statement}
We have taken several steps to ensure the reproducibility of our results. 
Implementation details for SCOPED are provided in \cref{app:walkthrough}, with full descriptions of noise level selection in \cref{sec:noise_selection}, score computation in \cref{app:diffusion_background}, and sign correction in \cref{app:sign}. 
Additional ablations, including single-step versus two-step variants and noise-level sensitivity, are reported in \cref{sec:ablation_1step} and \cref{sec:ablation_2step}. 
Vision experiments use the open-source CelebA DDPM checkpoint from \citet{heng2024out}, RL experiments rely on the standard D4RL datasets and our own trained diffusion models. 
We will release code for the vision experiments to ensure easy replication.

\section{Additional Background on Diffusion Models}
\label{app:diffusion_background}

Diffusion models \citep{sohl2015deep, ho2020denoising, song2021scorebased, karras2022elucidating} are generative models that learn to invert a forward noising process. Starting from clean data $x_0 \sim p_0(x)$, the forward process gradually perturbs samples toward Gaussian noise, producing intermediate marginals $p_t(x)$. A neural network is trained to approximate the score function $s_\theta(x_t,t) \approx \nabla_x \log p_t(x_t)$ across noise levels. Sampling then integrates a reverse-time differential equation, expressed either as a stochastic differential equation (SDE) or its deterministic probability flow ordinary differential equation (ODE), to transform noise back into data. Intuitively, high-frequency details disappear at small amounts of noise, while coarse or global structure is destroyed at large noise levels \citep{ho2020denoising, song2021scorebased}.

\paragraph{DDPM.}
The Denoising Diffusion Probabilistic Model (DDPM) \citep{ho2020denoising} defines a discrete forward Markov chain with variance schedule $\{\beta_t\}_{t=1}^T$. Let $\alpha_t = 1 - \beta_t$ and $\bar\alpha_t = \prod_{i=1}^t \alpha_i$, and define the effective noise scale $\sigma_t^2 = 1 - \bar\alpha_t$. The step-$t$ corruption of the clean data is
\[
x_t \,=\, \sqrt{\bar\alpha_t}\,x_0 \,+\, \sigma_t\,\epsilon, \qquad \epsilon \sim \mathcal N(0,I),
\]
so that $p(x_t \mid x_0) = \mathcal N(\sqrt{\bar\alpha_t}\,x_0,\, \sigma_t^2 I)$. A network $\epsilon_\theta(x_t,t)$ is trained to predict $\epsilon$ via
\[
\mathcal L_{\mathrm{DDPM}}(\theta) \;=\; \mathbb E_{t,x_0,\epsilon}\big[\|\epsilon_\theta(x_t,t) - \epsilon\|^2\big].
\]
Using $\nabla_x \log p_t(x_t) = -\,\mathbb E[\epsilon \mid x_t]/\sigma_t$, define $s_\theta(x_t,t) := -\,\epsilon_\theta(x_t,t)/\sigma_t$. Then
\[
\mathcal L_{\mathrm{DDPM}}(\theta)
= \mathbb E_{t,x_0,\epsilon}\Big[\sigma_t^2\,\big\|\,s_\theta(x_t,t) + \tfrac{\epsilon}{\sigma_t}\big\|^2\Big],
\]
which is a denoising score matching objective with noise level weight $w(t) = \sigma_t^2$, where $w(\cdot)$ controls the relative importance of different noise scales.

\paragraph{Continuous SDE formulation.}
Another general view of diffusion models specifies the forward process as an SDE \citep{song2021scorebased}. Here $f(x_t,t)$ is a drift term, $g(t)$ is the diffusion coefficient, and $B_t$ denotes standard Brownian motion (with $\bar B_t$ its time-reversed counterpart). These functions specify the forward noising process whose marginals are $p_t(x_t)$:
\[
dx \,=\, f(x_t,t)\,dt \,+\, g(t)\,dB_t, \qquad x_0 \sim p_0(x),
\]
with reverse SDE
\[
dx \,=\, \big[f(x_t,t) - g(t)^2 \nabla_x \log p_t(x_t)\big]dt \,+\, g(t)\,d\bar B_t,
\]
and equivalent probability flow ODE
\[
\frac{dx}{dt} \,=\, f(x_t,t) - \tfrac{1}{2} g(t)^2 \nabla_x \log p_t(x_t).
\]
When viewed in this way, training reduces to the denoising score matching (DSM) objective for continuous noise scales, arising from a process similar to DDPM. Concretely, letting $z_\sigma := x_0 + \sigma \epsilon$ with $\epsilon \sim \mathcal N(0,I)$ and $\sigma > 0$, the score estimator $s_\theta(z_\sigma,\sigma)$ is trained with the loss
\[
\mathcal L_{\mathrm{DSM}}(\theta)
= \mathbb E_{x_0,\sigma,\epsilon}\Big[w(\sigma)\,\big\|\,s_\theta(z_\sigma,\sigma) + \tfrac{\epsilon}{\sigma}\big\|^2\Big].
\]
Here $w(\sigma)\!\ge\!0$ is the general weighting function across noise scales and captures parameterization or preconditioning of the model. In the DDPM setting, taking $\sigma \in \{\sigma_t\}_{t=1}^T$ and predicting noise recovers $w(t)=\sigma_t^2$ as in DDPM.

\paragraph{EDM.}
One concrete instantiation of the continuous SDE framework is Elucidated Diffusion Models (EDM) \citep{karras2022elucidating}, which provides a principled continuous noise schedule $\sigma(t)$, input and output preconditioning, and solver choices that improve stability and sample quality. EDM often parameterizes the score function in terms of a denoiser $D_\theta(x,\sigma)$ via the relation \[
s_\theta(x,\sigma) \;=\; \frac{D_\theta(x,\sigma) - x}{\sigma^2}.
\]

\paragraph{Connection.}
DDPM and EDM are both instantiations of the same score-based framework: DDPM corresponds to a discrete, variance-preserving SDE with $w(t)=\sigma_t^2$, while EDM uses a continuous noise scale and preconditioning that imply a different effective weighting. Both ultimately train a network to approximate the score $\nabla_x \log p_t(x_t)$, providing a common foundation that our method exploits for OOD detection and highlighting the generality of our approach across diffusion model variants.

\section{Sign Ambiguity in the Typicality Ratio}
\label{app:sign}

The typicality ratio
\[
T(x) \;=\; \frac{\|s_\theta(x_t,t)\|^2}{-\mathrm{Tr}(\nabla_x s_\theta(x_t,t))}
\]
admits a global sign ambiguity. The numerator is a squared norm, which removes orientation, while the denominator involves the Jacobian trace without any loss of sign information.

To enforce consistency, we multiply $T(x)$ by a global sign factor derived from the score itself: 
\[
\text{sign} \;=\; \mathrm{sign}\!\left(\sum_i s_\theta(x_t,t)_i\right).
\]
Our implementation applies this factor to all OOD scores. We show in \cref{sec:signambig} that this correction is important for vision tasks.

\section{Signal-to-Noise Ratio of CIFAR-10}
\label{app:snr_cifar10}

We define the fraction of signal at timestep $t$ as 
\[
\frac{E_{\text{clean}}}{E_{\text{clean}} + E_{\text{noise}}},
\] 
where $E_{\text{clean}}$ is the average energy of clean images from the dataset and 
$E_{\text{noise}}$ is the energy contributed by Gaussian noise at timestep $t$. 
This quantity can be viewed as a normalized signal-to-noise ratio: it measures the proportion of 
total energy attributable to signal rather than noise, and is bounded between 0 and 1. 
This fraction decreases steadily with timestep, with early timesteps preserving fine detail and later ones becoming noise-dominated. All quantities are averaged over CIFAR-10.

\begin{figure}[H]
\centering
\includegraphics[width=0.6\textwidth]{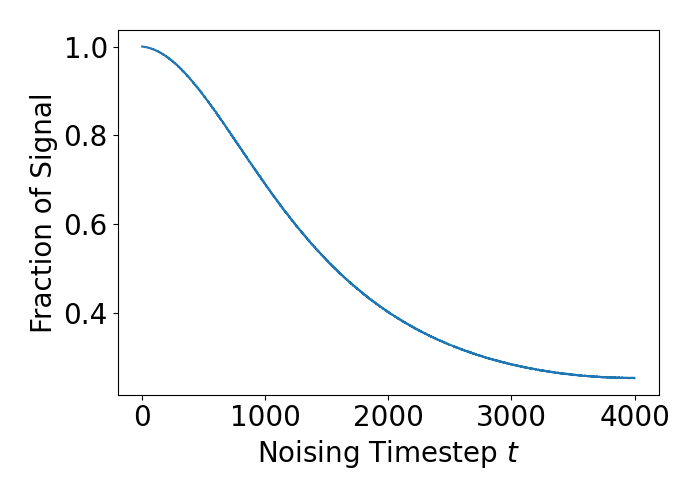}
\caption{\textbf{Signal-to-noise ratio curve for CIFAR-10.} 
The fraction of signal retained under the forward diffusion process decays steadily with timestep. Early timesteps preserve fine image detail, while later ones become noise-dominated.}
\label{fig:cifar10_snr}
\end{figure}

\section{Implementation Walkthrough (Vision Experiments)}
\label{app:walkthrough}

We summarize here the procedure for computing SCOPED scores in the vision experiments.

\begin{enumerate}
    \item \textbf{Backbone model.} We use the publicly released unconditional DDPM pretrained on CelebA by \citet{heng2024out} (available \href{https://huggingface.co/ajrheng/diffpath/tree/main}{here}) as our diffusion backbone. This ensures fair comparison to recent prior diffusion-based OOD results like \citep{heng2024out}. 
    \item \textbf{Select noise levels.} For vision tasks, we evaluate at two timesteps: $t=1$ (fine detail preserved) and $t=300$ (mid-level, $\sim$95\% signal retained), chosen from the signal-to-noise ratio curve (\cref{app:snr_cifar10}).
    \item \textbf{Offline calibration.} Compute score and curvature values $T(x)$ for in-distribution (ID) data at each selected noise level, and fit a kernel density estimate (KDE) to these values.
    \item \textbf{Test-time scoring.} For each candidate input $x_0$, corrupt to $x_t$ at the chosen timestep, run the model to obtain the score $s_\theta(x_t,t)$, and estimate curvature $\kappa(x_t) = \mathrm{Tr}(\nabla_x s_\theta(x_t,t))$ using Hutchinson’s trace estimator.
    \item \textbf{Form the statistic.} Compute
    \[
    T(x) \;=\; \frac{\|s_\theta(x_t,t)\|^2}{-\kappa(x_t) + \varepsilon}\,,
    \]
    and apply a global sign factor as described in \cref{app:sign}, where $\varepsilon > 0$ is a small constant added for numerical stability to avoid division by zero.
    \item \textbf{Aggregate across timesteps.} For each timestep, evaluate $T(x)$ under its ID KDE and compute the negative log-likelihood (NLL).  
        \begin{enumerate}
            \item In the two-step variant, take the maximum NLL across the two preselected timesteps as the anomaly score.  
            \item In the single-step variant, use the NLL at a single preselected timestep only.  
            \item In the oracle variant, select the single timestep that yields the highest AUROC with hindsight.
        \end{enumerate}
    \item \textbf{Thresholding.} For evaluation we report AUROC (threshold-free). 
    For deployment, a cutoff can be fixed from ID data alone, e.g., the $(1-\alpha)$ quantile of the 
    in-distribution score distribution, where $\alpha$ is the desired false positive rate 
    (e.g., $\alpha = 0.05$ corresponds to a 5\% ID false alarm rate).

\end{enumerate}

\section{Offline DMC Results}

We provide results for SCOPED on offline DeepMind Control Suite (DMC) replay buffers 
collected by deploying SAC policies to proficiency in various tasks. These experiments test whether SCOPED can distinguish distribution shifts induced by either different reward functions (e.g., \texttt{reacher-easy} vs.\ \texttt{reacher-hard}, \texttt{humanoid-stand} vs. \texttt{humanoid-walk}) or by independently trained policies within the same task. As shown in \cref{tab:dmc}, SCOPED achieves perfect separation (AUROC = 1.0) in all cases.

\label{app:offlinedmc}
\begin{table}[H]
\centering
\caption{SCOPED AUROC scores under DMC environment shifts.}
\label{tab:dmc}
\begin{tabular}{lcc}
\toprule
Case & Train buffer → Eval buffer & AUROC (OOD) \\
\midrule
Reacher & easy $\rightarrow$ hard & 1.000 \\
                       & hard $\rightarrow$ easy & 1.000 \\
Finger-Turn    & easy $\rightarrow$ hard & 1.000 \\
                       & hard $\rightarrow$ easy & 1.000 \\
Humanoid  & stand $\rightarrow$ walk & 1.000 \\
                       & walk $\rightarrow$ stand & 1.000 \\
Reacher hard policy realizations     & seedA $\rightarrow$ seedB & 1.000 \\
                       & seedB $\rightarrow$ seedA & 1.000 \\
Finger hard policy realizations      & seedA $\rightarrow$ seedB & 1.000 \\
                       & seedB $\rightarrow$ seedA & 1.000 \\
\bottomrule
\end{tabular}
\end{table}

\section{Ablation Over Single Step Choices in Noise Schedule for SCOPED}
\label{sec:ablation_1step}
We ablate the choice of timestep when SCOPED is deployed at a single noise level. \cref{tab:onestep} reports AUROC across vision dataset pairs for individual timesteps as well as an oracle baseline that selects the best-performing timestep with hindsight. Average AUROC is highest at mid-level timesteps, though performance at individual steps still varies across dataset pairs. This bolsters our claim, based on the SNR study, that combining early and mid-level evaluations yields a more robust detector than relying on a single fixed step.

\begin{table}[H]
\begin{center}
\scriptsize
\caption{AUROC when SCOPED is deployed at individual timesteps in the noise schedule. Bold entries indicate the best value in each column. The SCOPED (Oracle) row assumes a priori knowledge of both in-distribution and out-of-distribution data and selects the timestep in the denoising schedule that maximizes AUROC.}
\label{tab:onestep}
\begin{tabular}{lcccccccccc}
\toprule
\textbf{Timestep} & \multicolumn{3}{c}{\textbf{C10 vs}} & \multicolumn{3}{c}{\textbf{SVHN vs}} & \multicolumn{3}{c}{\textbf{CelebA vs}} & \textbf{Avg} \\
 & SVHN & CelebA & C100 & C10 & CelebA & C100 & C10 & SVHN & C100 & \\
\cmidrule(lr){2-4}\cmidrule(lr){5-7}\cmidrule(lr){8-10}
1   & 0.6671 & \textbf{0.8852} & \textbf{0.7678} & 0.8058 & 0.8773 & 0.6534 & 0.9164 & 0.8891 & 0.7247 & 0.7985 \\
2   & 0.6189 & 0.8620 & 0.6947 & 0.8057 & 0.9843 & 0.8844 & 0.8396 & 0.9663 & 0.7267 & 0.8203 \\
3   & 0.7241 & 0.7346 & 0.5570 & 0.8830 & 0.9845 & 0.8913 & 0.5437 & 0.8689 & 0.5601 & 0.7497 \\
4   & 0.7812 & 0.6241 & 0.5459 & 0.9081 & 0.9816 & 0.9053 & 0.6319 & 0.9474 & 0.6425 & 0.7742 \\
5   & 0.8031 & 0.5793 & 0.5465 & 0.9188 & 0.9804 & 0.9159 & 0.6257 & 0.9630 & 0.6472 & 0.7755 \\
100 & 0.9118 & 0.3645 & 0.5448 & 0.9623 & 0.9887 & 0.9540 & 0.6357 & 0.9797 & 0.6790 & 0.7801 \\
200 & \textbf{0.9638} & 0.2801 & 0.7555 & \textbf{0.9925} & \textbf{1.0000} & \textbf{0.9752} & 0.7408 & 0.9999 & 0.9384 & 0.8496 \\
300 & 0.7744 & 0.8672 & 0.4596 & 0.9757 & \textbf{1.0000} & 0.9655 & \textbf{0.9383} & \textbf{1.0000} & \textbf{0.9705} & 0.8835 \\
400 & 0.7827 & 0.8585 & 0.4633 & 0.9275 & 0.9957 & 0.9118 & 0.9140 & 0.9939 & 0.9644 & 0.8680 \\
500 & 0.8585 & 0.6506 & 0.4848 & 0.9235 & 0.9376 & 0.9357 & 0.8144 & 0.9628 & 0.8773 & 0.8272 \\
\midrule
SCOPED (Oracle) & 0.9638 & 0.8852 & 0.7678 & 0.9925 & 1.0000 & 0.9752 & 0.9383 & 1.0000 & 0.9705 & 0.9437 \\
\bottomrule
\end{tabular}
\end{center}
\end{table}

\section{Ablation Over Two Step Choices in Noise Schedule for SCOPED}
\label{sec:ablation_2step}

We ablate the choice of timestep pairs when SCOPED is deployed with two evaluations. \cref{tab:twostep} reports AUROC across vision dataset pairs for combinations of one early (low-noise) and one mid-level timestep. Average AUROC is consistently high across different choices, indicating that the two-step variant is robust to the precise placement of timesteps and does not depend on careful tuning.

\begin{table}[H]
\begin{center}
\scriptsize
\caption{
Ablation of timestep selection for SCOPED with NFE=2. Each row shows AUROC when pairing one early (low-noise) timestep with one mid-noise timestep from the diffusion schedule. Results are reported across three ID vs OOD benchmarks, with averages in the final column. Performance remains consistently high across different choices, indicating robustness to timestep selection.
}
\label{tab:twostep}
\begin{tabular}{lccccccccccc}
\toprule
\textbf{Time Steps} & \multicolumn{3}{c}{\textbf{C10 vs}} & \multicolumn{3}{c}{\textbf{SVHN vs}} & \multicolumn{3}{c}{\textbf{CelebA vs}} & \textbf{Avg} \\
 & SVHN & CelebA & C100 & C10 & CelebA & C100 & C10 & SVHN & C100 & \\
\cmidrule(lr){2-4} \cmidrule(lr){5-7} \cmidrule(lr){8-10}
(1,100) & 0.904 & 0.367 & 0.550 & 0.960 & 0.989 & 0.944 & 0.640 & 0.975 & 0.676 & 0.778 \\
(1,200) & 0.904 & 0.285 & 0.812 & 0.987 & 0.996 & 0.967 & 0.735 & 0.992 & 0.922 & 0.844 \\
(1,300) & 0.814 & 0.940 & 0.468 & 0.973 & 0.996 & 0.959 & 0.924 & 0.995 & 0.961 & 0.892 \\
(1,400) & 0.793 & 0.926 & 0.470 & 0.924 & 0.994 & 0.907 & 0.906 & 0.991 & 0.954 & 0.874 \\
(1,500) & 0.866 & 0.654 & 0.483 & 0.926 & 0.937 & 0.929 & 0.824 & 0.963 & 0.872 & 0.828 \\
(2,100) & 0.862 & 0.376 & 0.554 & 0.948 & 0.996 & 0.952 & 0.639 & 0.980 & 0.680 & 0.776 \\
(2,200) & 0.802 & 0.346 & 0.761 & 0.976 & 0.999 & 0.969 & 0.725 & 0.997 & 0.927 & 0.834 \\
(2,300) & 0.839 & 0.936 & 0.486 & 0.965 & 0.999 & 0.961 & 0.922 & 0.998 & 0.969 & 0.897 \\
(2,400) & 0.826 & 0.939 & 0.480 & 0.912 & 0.997 & 0.911 & 0.919 & 0.994 & 0.965 & 0.883 \\
(2,500) & 0.810 & 0.773 & 0.521 & 0.905 & 0.957 & 0.931 & 0.823 & 0.965 & 0.881 & 0.841 \\
(3,100) & 0.875 & 0.391 & 0.546 & 0.956 & 0.995 & 0.950 & 0.633 & 0.974 & 0.674 & 0.777 \\
(3,200) & 0.816 & 0.516 & 0.618 & 0.980 & 0.998 & 0.968 & 0.709 & 0.992 & 0.910 & 0.834 \\
(3,300) & 0.860 & 0.892 & 0.485 & 0.969 & 0.999 & 0.960 & 0.897 & 0.994 & 0.947 & 0.889 \\
(3,400) & 0.845 & 0.896 & 0.476 & 0.928 & 0.996 & 0.916 & 0.897 & 0.990 & 0.946 & 0.877 \\
(3,500) & 0.819 & 0.745 & 0.518 & 0.925 & 0.967 & 0.931 & 0.806 & 0.963 & 0.860 & 0.837 \\
(4,100) & 0.891 & 0.395 & 0.545 & 0.960 & 0.995 & 0.953 & 0.627 & 0.978 & 0.665 & 0.779 \\
(4,200) & 0.858 & 0.509 & 0.595 & 0.984 & 0.998 & 0.970 & 0.702 & 0.993 & 0.882 & 0.832 \\
(4,300) & 0.877 & 0.840 & 0.490 & 0.971 & 0.998 & 0.962 & 0.888 & 0.995 & 0.931 & 0.884 \\
(4,400) & 0.864 & 0.853 & 0.479 & 0.934 & 0.996 & 0.919 & 0.876 & 0.992 & 0.928 & 0.871 \\
(4,500) & 0.851 & 0.681 & 0.520 & 0.933 & 0.968 & 0.940 & 0.801 & 0.967 & 0.847 & 0.834 \\
(5,100) & 0.897 & 0.398 & 0.545 & 0.963 & 0.995 & 0.955 & 0.624 & 0.978 & 0.661 & 0.780 \\
(5,200) & 0.871 & 0.503 & 0.587 & 0.985 & 0.998 & 0.969 & 0.706 & 0.993 & 0.871 & 0.831 \\
(5,300) & 0.882 & 0.815 & 0.487 & 0.972 & 0.998 & 0.963 & 0.868 & 0.995 & 0.919 & 0.878 \\
(5,400) & 0.871 & 0.829 & 0.482 & 0.936 & 0.996 & 0.921 & 0.878 & 0.992 & 0.919 & 0.869 \\
(5,500) & 0.863 & 0.638 & 0.525 & 0.938 & 0.968 & 0.945 & 0.783 & 0.969 & 0.846 & 0.830 \\

\midrule
\textbf{AVG} & \textbf{0.854} & \textbf{0.658} & \textbf{0.539} & \textbf{0.952} & \textbf{0.989} & \textbf{0.946} & \textbf{0.790} & \textbf{0.985} & \textbf{0.865} & \textbf{0.842} \\
\bottomrule
\end{tabular}
\end{center}
\end{table}

\section{Ablation Over Two Step Choices in Noise Schedule for SCOPED without Sign Ambiguity Term}
\label{sec:signambig}

We repeat the two-step ablation without applying the global sign correction described in \cref{app:sign}. \cref{tab:nosign} shows that while AUROC remains non-trivial when SCOPED is applied without a sign correction, average performance drops to 0.81 compared to 0.84. This confirms that the sign term is necessary for stable performance in vision benchmarks, even though raw score–curvature ratios can still capture some separation.

\begin{table}[H]
\begin{center}
\scriptsize
\caption{
Ablation of timestep selection for SCOPED with NFE=2 and no sign term. Each row shows AUROC when pairing one early (low-noise) timestep with one mid-noise timestep from the diffusion schedule. Results are reported across three ID vs OOD benchmarks, with averages in the final column. Performance remains consistently high across different choices, indicating robustness to timestep selection.
}
\label{tab:nosign}
\begin{tabular}{lcccccccccc}
\toprule
\textbf{Time Steps} & \multicolumn{3}{c}{\textbf{C10 vs}} & \multicolumn{3}{c}{\textbf{SVHN vs}} & \multicolumn{3}{c}{\textbf{CelebA vs}} & \textbf{Avg} \\
 & SVHN & CelebA & C100 & C10 & CelebA & C100 & C10 & SVHN & C100 & \\
\cmidrule(lr){2-4} \cmidrule(lr){5-7} \cmidrule(lr){8-10}
(1,100) & 0.863 & 0.367 & 0.543 & 0.963 & 0.994 & 0.955 & 0.642 & 0.983 & 0.685 & 0.777 \\
(1,200) & 0.752 & 0.280 & 0.408 & 0.996 & 0.999 & 0.996 & 0.617 & 0.992 & 0.656 & 0.744 \\
(1,300) & 0.850 & 0.899 & 0.863 & 0.996 & 0.999 & 0.997 & 0.602 & 0.987 & 0.634 & 0.870 \\
(1,400) & 0.855 & 0.891 & 0.861 & 0.987 & 0.997 & 0.986 & 0.590 & 0.973 & 0.615 & 0.862 \\
(1,500) & 0.884 & 0.651 & 0.699 & 0.864 & 0.940 & 0.862 & 0.584 & 0.950 & 0.606 & 0.782 \\
(2,100) & 0.858 & 0.374 & 0.542 & 0.947 & 0.993 & 0.943 & 0.639 & 0.980 & 0.682 & 0.773 \\
(2,200) & 0.781 & 0.334 & 0.452 & 0.980 & 0.997 & 0.982 & 0.612 & 0.989 & 0.651 & 0.753 \\
(2,300) & 0.876 & 0.890 & 0.838 & 0.983 & 0.997 & 0.984 & 0.599 & 0.984 & 0.630 & 0.865 \\
(2,400) & 0.889 & 0.899 & 0.848 & 0.955 & 0.993 & 0.958 & 0.585 & 0.970 & 0.614 & 0.857 \\
(2,500) & 0.829 & 0.737 & 0.697 & 0.852 & 0.953 & 0.853 & 0.581 & 0.948 & 0.603 & 0.784 \\
(3,100) & 0.875 & 0.389 & 0.543 & 0.956 & 0.995 & 0.949 & 0.633 & 0.974 & 0.673 & 0.776 \\
(3,200) & 0.797 & 0.500 & 0.502 & 0.985 & 0.998 & 0.985 & 0.608 & 0.985 & 0.646 & 0.778 \\
(3,300) & 0.903 & 0.869 & 0.814 & 0.987 & 0.998 & 0.987 & 0.594 & 0.982 & 0.628 & 0.863 \\
(3,400) & 0.909 & 0.874 & 0.821 & 0.970 & 0.995 & 0.970 & 0.586 & 0.969 & 0.611 & 0.856 \\
(3,500) & 0.835 & 0.715 & 0.626 & 0.878 & 0.965 & 0.875 & 0.580 & 0.947 & 0.600 & 0.780 \\
(4,100) & 0.891 & 0.395 & 0.545 & 0.960 & 0.995 & 0.953 & 0.627 & 0.978 & 0.665 & 0.779 \\
(4,200) & 0.836 & 0.506 & 0.513 & 0.988 & 0.998 & 0.988 & 0.606 & 0.987 & 0.646 & 0.785 \\
(4,300) & 0.923 & 0.840 & 0.801 & 0.990 & 0.998 & 0.989 & 0.597 & 0.985 & 0.628 & 0.861 \\
(4,400) & 0.929 & 0.850 & 0.813 & 0.976 & 0.996 & 0.976 & 0.585 & 0.973 & 0.611 & 0.857 \\
(4,500) & 0.866 & 0.676 & 0.611 & 0.888 & 0.968 & 0.883 & 0.583 & 0.953 & 0.602 & 0.781 \\
(5,100) & 0.897 & 0.398 & 0.545 & 0.963 & 0.995 & 0.955 & 0.624 & 0.977 & 0.661 & 0.779 \\
(5,200) & 0.847 & 0.502 & 0.514 & 0.990 & 0.998 & 0.989 & 0.606 & 0.987 & 0.646 & 0.787 \\
(5,300) & 0.929 & 0.815 & 0.793 & 0.991 & 0.998 & 0.991 & 0.595 & 0.985 & 0.626 & 0.858 \\
(5,400) & 0.935 & 0.829 & 0.809 & 0.979 & 0.996 & 0.978 & 0.585 & 0.974 & 0.614 & 0.855 \\
(5,500) & 0.878 & 0.638 & 0.600 & 0.891 & 0.968 & 0.887 & 0.581 & 0.954 & 0.603 & 0.778 \\
\midrule
\textbf{AVG} & \textbf{0.867} & \textbf{0.645} & \textbf{0.664} & \textbf{0.957} & \textbf{0.989} & \textbf{0.955} & \textbf{0.602} & \textbf{0.975} & \textbf{0.633} & \textbf{0.810} \\
\bottomrule
\end{tabular}
\end{center}
\end{table}

\end{document}